\documentclass[11pt,a4paper]{article}
\usepackage[final]{emnlp2021}
\usepackage{times}
\usepackage{latexsym}
\usepackage{enumitem}
\usepackage{booktabs}
\usepackage{array}
\usepackage{url}
\usepackage{tikz}
\usepackage{amsmath}
\usepackage{mathtools}
\usepackage{cleveref}
\usepackage[normalem]{ulem}
\graphicspath{{img/}}

\newcommand{\footnotehref}[2]{\footnote{\href{#1}{#2}}}
\definecolor{darkgreen}{rgb}{0.2,0.7,0.2}
\definecolor{darkred}{rgb}{0.7,0.2,0.2}

\def\blockfirst#1#2#3#4#5{
    \vspace{0.07cm}
    \hspace{-0.34cm}
    \begin{tikzpicture}[x=0.7em, y=1.5em]
        \draw [use as bounding box] (0,0) rectangle (5,1);
        \fill [fill=black] (0,0) rectangle (1,#1/16);
        \fill [fill=black] (1,0) rectangle (2,#2/16);
        \fill [fill=black] (2,0) rectangle (3,#3/16);
        \fill [fill=black] (3,0) rectangle (4,#4/16);
        \fill [fill=black] (4,0) rectangle (5,#5/16);
    \end{tikzpicture}
    \hspace{-0.3cm}
}

\def\blocksimple#1{
    \vspace{-0.014cm}
    \hspace{-0.14cm}
    \begin{tikzpicture}[x=1.45em, y=1.4em]
    \draw [use as bounding box] (0,0) rectangle (1,1);
    \fill [fill=black] (0,0) rectangle (1, #1);
    \end{tikzpicture}
    \hspace{-0.14cm}
}

    \title{Neural Machine Translation Quality and Post-Editing Performance}
\author{
    Vilém Zouhar\textnormal{,}$^\star$ Aleš Tamchyna\textnormal{,}$^\dagger$ Martin Popel\textnormal{,}$^\star$ Ondřej Bojar$^\star$ \\ \AND
    Charles University, Faculty of Mathematics and Physics$^\star$ \\
    Institute of Formal and Applied Linguistics \\
    Malostranské Náměstí 25, 118 00 Prague, Czech Republic \\
    {\tt \{zouhar, popel, bojar\}@ufal.mff.cuni.cz} \\ \AND
    Memsource$^\dagger$ \\
    Spálená 108/51, Prague, Czech Republic \\
    {\tt ales.tamchyna@memsource.com}
}

\date{}

\begin{document}
\maketitle
\begin{abstract}

    We test the natural expectation that using MT in professional translation saves human processing time. The last such study was carried out by \citet{sanchez2016machine} with phrase-based MT, artificially reducing the translation quality. In contrast, we focus on neural MT (NMT) of high quality, which has become the state-of-the-art approach since then and also got adopted by most translation companies.
        
    Through an experimental study involving over 30 professional translators for English$\rightarrow$Czech translation, we examine the relationship between NMT performance and post-editing time and quality. Across all models, we found that better MT systems indeed lead to fewer changes in the sentences in this industry setting. The relation between system quality and post-editing time is however not straightforward and, contrary to the results on phrase-based MT, BLEU is definitely not a stable predictor of the time or final output quality.

\end{abstract}

\section{Introduction}

Machine translation is increasingly utilized in the translation and localization industry. One of the most common use cases is MT post-editing where human translators use MT outputs as a starting point and make edits to obtain the final translation. This process has been shown to be more efficient than translating from scratch, i.e. it and can lead to reductions in cost and delivery time \citep{plitt2010productivity}.

While various studies have looked at how MT post-editing affects translation quality and speed in general, few have attempted to measure how the \textit{quality} of MT outputs affects the productivity of translators. In this work, we attempt to answer the following questions:

\begin{itemize}[]
    \item How strong is the relationship between MT quality and post-editing speed?
    \item Does MT quality have a measurable impact on the quality of the post-edited translation?
    \item Is the effect of MT quality still persistent in a second round of post-editing (``revision'')?
    \item Is the post-editing process different when human translation is used instead of MT as the input?
    \item How large are the edits in the different rounds of post-editing?
\end{itemize}

We have carried out a large-scale study on one language pair that involved over 30 professional translators and translation reviewers, who worked in two stages to post-edit outputs of 13 different sources (11 MT engines, 1 raw source, 1 human reference).
This allowed us to collect not only the post-editing times but also to estimate the quality of the produced results.
Based on this data, we present an in-depth analysis and provide observations and recommendations for utilizing MT in localization workflows.
While the task for humans for both rounds is the same (improve a given translation by freely editing it), we strictly distinguish the first and second rounds, using the term ``post-editor'' in the first round and ``reviewer'' in the second round.
% The reviewers thus always operate on the outputs of our post-editors.
% Zakomentované protože není pravda. -V.

We make the code along with all collected data (including all translations) publicly available.\footnotehref{https://github.com/ufal/nmt-pe-effects-2021}{github.com/ufal/nmt-pe-effects-2021}

\section{Related Work}

One of the earliest experiments that noticed a significant correlation between various automatic evaluation metrics and post-editing speed was performed by \citet{tatsumi2009correlation}.
The survey of \citet{koponen2016machine} briefly covers the history of post-editing and pinpoints the two main topics: effort and final output quality. The authors conclude that post-editing improves both compared to translating the original text from scratch, given suitable conditions (good MT quality and translator experience with post-editing).
Experiments were done by \citet{mitchell2013community} and \citet{koponen2015correctness} show that purely monolingual post-editing leads to results of worse quality than when having access to the original text as well.
Finally, \citet{koponen2013translation} comments on the high variance of post-editors, which is a common problem in post-editing research  \citep{koponen2016machine}.

Interactive MT is an alternative use case of computer-assisted translation and it is possible that effort or behavioural patterns in interactive MT could be used as a different proxy extrinsic measure for MT quality.
Post-editor productivity has also been measured in contrast to interactive translation prediction by \citet{sanchis2014interactive}.

\paragraph{Similar Experiments.}

Our work is most similar to \citet{sanchez2016machine} and \citet{koehn2014impact}, which served as a methodological basis for our study.
% Our work is, however, different in several key aspects.

While most of the previous works focused on Statistical MT, we experiment solely with Neural MT (NMT) models. Many studies have shown that NMT has very different properties than older MT models when it comes to post-editing \citep{koponen2019product}. For instance, NMT outputs tend to be very fluent which can make post-editing more cognitively demanding and error-prone as suggested by \citet{castilho2017neural}. \citet{popel-et-al:2020} showed that in a specific setting, the adequacy of NMT is higher than that of human translators. 
We believe that the relationship between NMT system quality and PE effort is not a simple one and that older results based on statistical MT may not directly carry over to NMT. The first of the six challenges listed by \citet{koehn2017six} suggests that fluency over adequacy can be a critical issue: \textit{NMT systems have lower quality out of domain, to the point that they completely sacrifice adequacy for the sake of fluency.} 

Additionally, our focus is state-of-the-art NMT systems, which was not true for \citet{sanchez2016machine}, who constructed 9 artificially severely degraded statistical phrase-based MT systems.
The experiment by \citet{koehn2014impact} used only 4 MT systems. Our focus is motivated by the industry's direct application:
\textit{Considering the cost of skilled staff and model training, what are the practical benefits of improving MT performance?}

In contrast to the previous setups, we evaluate two additional settings: post-editing human reference and translating from scratch, corresponding to a theoretical\footnote{In fact, humans never produce the same translation, so BLEU of 100 is unattainable, and the source text often contains some tokens appearing also in the output, so not translating can reach BLEU scores of e.g. 3 or 4.} BLEU of 100 and 0, respectively.
We also consider the quality of the PE output and not only the process itself.

\citet{sanchez2016machine} found a linear relationship between BLEU and PE effort: \textit{for each 1-point increase in BLEU, there is a PE time decrease of 0.16 seconds per word, about 3-4\%.} The performance of the MT systems they use is, however, close to uniformly distributed between $24.85$ and $30.37$.
The observed linear relationship can then be partially attributed to the lower MT performance of artificially uniformly distributed MT systems.
% \MP{Vysvětlit rozpor mezi výsledky Sanchez-Torron a našimi by bylo skvělé.
% Jenže M1--M9 jsou taky uměle underperforming kratším trénováním (byť na plných datech) a poměry synth:auth.
% Ty jejich systémy jsou trénovany vždy na polovině dat.
% Jejich nejhorší systém má 81 \% BLEU toho nejlepšího, u nás 68 \%,
% nelze tedy ani říct, že by měli větší rozsah BLEU.
% Možná mají lepší rovnoměrnost pokrytí toho rozsahu, ale jen o trochu.
% }
% \OB{To shrnuti Sancheze-Torrona a ten popis jejich rozsahu systemu je zajimavy, dal bych ho vic do kontrastu s nasim rozsahem. Ale podobne jako Martin nevidim zatim zadny duvod, proc to vyslo jinak, cili to bych neoteviral, nerozebiral.}
% \OB{V komentari ve zdrojaku je zde muj napad, proc to Philippovi vyslo linearne a nam ne.
%> - linearni vztah mezi BLEU a casem mohl Philippovi vyjit, protoze lepsi a lepsi
%> PBMT dela lepsi a lepsi ngramy, coz BLEU vidi. Predstavme si, ze kazdy bod BLEU
%> znamena jedno dobre prelozene slovicko navic. Toto slovicko pak posteditorovi
%> pomuze, nemusi ho vymyslet a mozna ani psat. Odlisnosti v posteditacnim case
%> pro ruzne tezke vety, ruzne aktivni posteditory ap. se pak (zrejme) vyrovnaji
%> diky nahodnemu sumu.
%> U nas ocivodne neplati, ze lepsi BLEU je lepsi preklad, a tim vic neplati ta
%> zjednodusena predstava, ze kazdy bod BLEU znamena dobre slovicko navic.
% }

\paragraph{Neural MT.} An experiment by \citet{koponen2020mt} considers 4 neural MT systems in a similar setting. The quality of these systems is below the state of the art.\footnote{Document-level BLEU of $19.3$ on miscellaneous FI$\rightarrow$SV OPUS \citep{tiedemann2012parallel} data. Current state of the art is $29.5$ on the FIKSMÖ benchmark \citep{tiedemann2020fiskmo}.}
The focus of this work was also to measure the difference between translating from scratch and post-editing, which was confirmed to be in favour of the latter. 
The contrast of using translation memories and NMT on little-explored language pairs was examined by \citet{laubli2019post}.

\section{Experiment Design}

In this section, we thoroughly describe the design of the study, including the used data, MT engines and the translation process.

\subsection{Documents}

In total, we used $99$ source lines (segments) of $8$ different parallel English documents for which Czech human reference translations were available.
One line can contain more than one sentence, which is reflected by the rather high average sentence length of $25$ words.
% \OB{Alesi, prosim potvrd, ze ani Memsource to lidem nerozsekal na vety, tj. ze opravdu videli atypicky dlouhe segmenty. Ja spis tedy doufam, ze jim to rozsekal, takze pro lidi to bylo vic nez 99 textovych policek.}
% \AT{Nerozsekal, on by byl problem, co pak s temi MT vystupy, musely by se nejak nasekat taky. Cili ano, vety byly dlouhe.}
We chose the domains to mirror common use-cases in localization: $36$ lines of news texts (WMT19 News testset), $29$ lines from a lease agreement (legal text), $23$ lines from an audit document \cite{zouhar:markable}, and $11$ lines of technical documentation \cite{qtleap}.
The translators received all documents joined together in a single file, with clearly marked document boundaries. For clarity, we will refer to the whole set simply as \textit{``file''} and the individual parts as \textit{``documents''}.

\subsection{Machine Translation Models}

In total, we used 13 MT models of various quality.
Models M01--M11 are based on the setup, training procedure and data of \citet{popel:2020:WMT}.
We chose this particular approach because it has been reported to reach human translation quality \citep{popel-et-al:2020}.
For our purposes, we reproduce the training, stopping it at various stages of the training process.
All MT systems translate sentences in isolation, with the exception of M11, which is a document-level system (replicating CUNI-DocTransformer in \citet{popel:2020:WMT}).
Systems MT01--MT10 differ only in the number of training steps, which affects also the ratio of authentic- and synthetic- data checkpoints in the hourly checkpoint averaging \citep{popel-et-al:2020}:
the best dev-set BLEU was achieved with 6 authentic-data and 2 synthetic-data checkpoints,
but we include also models with other ratios (cf. column \textit{ACh} in \Cref{tab:mt_overview}).

In addition to the internal MT system variants, we also included outputs of commercially available models by Google\footnotehref{https://cloud.google.com/translate/}{cloud.google.com/translate} and Microsoft.\footnotehref{https://azure.microsoft.com/en-us/services/cognitive-services/translator/}{azure.microsoft.com/en-us/services/cognitive-services/translator/}

Overview of all the 13 MT systems is provided in \Cref{tab:mt_overview}. Although the range of BLEU scores is very large ($25.35$--$37.44$), the scores are not spread out evenly (average $34.65$).\footnote{
We also experimented with BERTScore but its Pearson correlation with BLEU is $0.9939$. This would lead to the same observations and conclusions.}
Most of the systems are concentrated in the upper half of the range. This better reflects realistic scenarios in localization workflows where users can typically decide among several engines of comparable but not identical performance.

\begin{table}[ht]
    \centering
    \begin{tabular}{lrrrc}
\toprule
Model & TER & BLEU & Steps [k] & ACh \\\midrule
M01   & 0.729 & 25.35 &   25.4 & 8\\
M02   & 0.678 & 31.61 &   29.0 & 8\\
M03   & 0.655 & 33.09 &   29.3 & 8\\
M04   & 0.648 & 33.63 &   33.0 & 8\\
M05   & 0.622 & 35.22 &   72.8 & 6\\ %6--7
M06   & 0.624 & 35.68 &  997.1 & 0\\
M07   & 0.604 & 36.58 & 1015.2 & 5\\
M08   & 0.600 & 36.41 & 1022.4 & 6\\ %6--7
M09   & 0.603 & 37.40 & 1055.0 & 8\\
M10   & 0.600 & 37.44 & 1058.6 & 6\\ %6--7
M11   & 0.601 & 37.37 &  698.5 & 5\\
Google & 0.623 & 37.56   & --  & --\\
Microsoft& 0.632 & 33.06 & --  & --\\
\bottomrule
\end{tabular}

    \caption{Overview of MT systems used.
        TER and BLEU were measured by SacreBLEU\footnotemark{} \citep{sacrebleu}.
        Steps mark the number of training steps in thousands.
        ACh is the number of authentic-data-trained checkpoints in an average of 8 checkpoints.
    }
    \label{tab:mt_overview}
\end{table}
\footnotetext{TER+t.tercom-nonorm-punct-noasian-uncased+v.1.5.1\newline BLEU+c.mixed+\#.1+s.exp+tok.13a+v.1.4.14}

\subsection{Translation Process}

We carried out the translation in two stages: MT post-editing stage and final revision stage.
For both stages, we used Memsource as the computer-assisted translation (CAT) tool.

\paragraph{(1) Post-editing} The documents were first translated by all 13 MT systems. In addition, we included a variant with no translation ("Source") and with a pre-existing reference translation ("Reference").\footnote{For simplicity, we refer to these two types of input (Reference and Source) also as MT systems.} The translated files were shuffled at document boundaries so that each document in the file was translated by a single MT system and no MT system appeared twice in a single file.
% Note that for some documents, no MT output was available (``Source") or the ``MT output'' was in fact a human translation (``Reference").

The resulting 15 files were given to 15 professional post-editors. Every post-editor worked with all $99$ lines. This stage provides the primary data for determining the amount of time the post-editors need to bring the candidates to the common industry standards.
The post-editors are well used to carrying out this task and are familiar with the CAT tool. In the translation editor, the MT outputs (incl. Reference, indistinguishable) were offered as 100\% TM matches. No other technical tools (MT, TM etc.) were allowed. The post-editors received instructions mentioning that the provided translation may be manual or automatic (or missing in the case of Source). They were also asked to take any necessary breaks only at document boundaries marked in the input file.

\paragraph{(2) Revision} After the first post-editing, the results were examined by 17 professional reviewers and further refined. None of the first-phase post-editors was included in the set of reviewers. Before submitting the data for the second stage, we further shuffled the translations on the document-level so that each reviewer received a random mix of documents produced by different post-editors.

In addition to the post-edited documents, we also included the pre-existing reference translation and the output of the document-level MT system (M11) without post-editing.

Again, ``revision'' is a standard task in the industry. The proposed translation is pre-filled in the output fields and reviewers modify it as necessary.
The instructions mentioned that the proposed translation may be the result of manual post-editing of MT, manual translation with the help of MT or unedited MT output but suggested to fix only true errors: wrong translation, inaccuracies, grammar or style errors. Individual translators' preferences were supposed to be avoided.

The main goal of this stage is to measure how the quality of MT for post-editing affects the quality of the final translations.
The standard practice of the industry is to focus on the output, providing the final quality check and the last necessary fixes, not any laborious error annotation. To complement purely quantitative measures (review time and string difference of the translation before and after the review), we asked the reviewers to also perform a basic variant of LQA ("Linguistic Quality Assurance") based on the MQM-DQF framework.\footnotehref{http://www.qt21.eu/wp-content/uploads/2015/11/QT21-D3-1.pdf}{qt21.eu/wp-content/uploads/2015/11/QT21-D3-1.pdf} The reviewers classified errors into three categories: accuracy (adequacy), fluency and other.
Every comment also had a severity attached (0-neutral, 1-minor, 2-major, 3-critical).
We report severity averaged per number of sentences.

\subsection{Post-Editors}

We examine in detail the composition of our set of post-editors from the first phase, not of the reviewers.
The experience in translating and post-editing of our post-editors, shown in \Cref{tab:experience}, was slightly lower than that of \citet{sanchez2016machine}.

\begin{table}[ht]
    \center
    \begin{tabular}{lccc}
\toprule
(years) & $<$ 5 & $<$ 10 & $\ge$ 10 \\
\midrule
Translation experience\hspace{0.0cm} & 5 & 2 & 9 \\
Post-Editor experience\hspace{0.0cm} & 8 & 7 & 0 \\
\bottomrule
\end{tabular}
    \caption{Post-editors' (first phase) years of translation and post-editing experience.}
    \label{tab:experience}
\end{table}

\paragraph{Questionnaire} All post-editors were asked to fill in a questionnaire, which is inspired by \citet{sanchez2016machine} but further extends it with questions regarding translators' sentiment on MT. The results are shown in \Cref{tab:questionnaire} and are comparable to the data collected by \citet{sanchez2016machine} with the exception of a slightly higher positive opinion of using MT. Namely, we see a higher agreement on \textit{MT helps to translate faster} and less disagreement on \textit{Prefer PE to editing human translation}.
For questionnaire results of reviewers please see \Cref{sec:questionnaire_p2}.

There is a clear preference for using even imprecise TM matches (85--94\%) over MT output.
This corresponds to the general tendency towards the opinion that post-editing is more laborious than using a TM, which is an interesting contrast to the preference for post-editing over translation from scratch.
The question about preferring to post-edit human over machine output shows a perfect Gaussian distribution, i.e. the lack of such preference in general.
We see some level of trust in MT in the process helping to improve translation consistency and produce overall better results.
% \AT{Je ta nasledujici veta potreba?} (This trust may not be much justified if one knows how the MT systems are constructed.)
%
For personal use, the post-editors know MT technology and use it for languages they do not speak.

\begin{table}[t]
    \center
    \begin{tabular}{>{\raggedright}p{5.7cm}m{1.15cm}}
\toprule
Question &\hspace{-0.35cm}Response\hspace{-0.2cm} \\
\midrule
Comfortable with post-editing human-like (perfect) quality & \blockfirst{1}{1}{7}{7}{0} \\
Comfortable with post-editing less-than-perfect quality & \blockfirst{0}{7}{2}{5}{2} \\
Prefer PE to translating from scratch (without a TM) & \blockfirst{1}{4}{4}{6}{1} \\
MT helps to maintain translation consistency & \blockfirst{0}{4}{4}{6}{2} \\
MT helps to translate faster & \blockfirst{0}{1}{2}{9}{4} \\
PE is more laborious than translating from scratch or with a TM & \vspace{0.4cm} \blockfirst{1}{2}{5}{7}{1} \\
Prefer PE to processing 85--94\% TM matches & \blockfirst{0}{7}{5}{3}{1} \\
Prefer PE to editing a human translation & \blockfirst{2}{4}{5}{3}{2} \\
% new questions
MT helps to produce better results & \blockfirst{1}{1}{6}{6}{2} \\
Often use MT outside of work for known languages & \blockfirst{1}{6}{4}{5}{0} \\
Often use MT outside of work for unknown languages & \blockfirst{0}{3}{4}{5}{4} \\
\bottomrule
\end{tabular}
    \caption{Post-editors' (first phase) answers regarding their profession on the Likert scale (leftmost bar = Strongly Disagree, rightmost bar = Strongly Agree), TM = translation memory.}
    \label{tab:questionnaire}
\end{table}

% OB zakomentovava, uz receno vyse
%\paragraph{Translator Environment}
%
%Post-editors of the first phase worked in their usual professional environment. They were asked not to use any external MT system and to take breaks only on document boundaries.

\section{Post-Editing Effort}

To measure the post-editing effort, we examine first the differences between provided translation and the post-edited output (Section~\ref{sec:edits}). We then focus on the time spent post-editing, which is an extrinsic evaluation of this task (Section~\ref{sec:time}).

\subsection{Edits}\label{sec:edits}

\paragraph{Output Similarity}

The post-edited outputs of MT systems had $21.77\pm0.11$ tokens per line, which is slightly higher than for the original candidates ($21.10\pm0.12$). Also, Reference ($21.76$ tokens per line) got in comparison to MT systems less long in post-editing, reaching $22.10$.\footnote{$95\%$ confidence interval based on Student's t-test.}

To measure the distance between the provided translations and the post-editors' output, we used character n-gram F-score (ChrF, \citealp{metrics_chrf}). For computation we again use SacreBLEU\footnote{ChrF2+numchars.6+space.false+version.1.4.14} \citep{sacrebleu}.

\Cref{tab:axp3} shows the measured ChrF similarities. For Source, the English input was used and received a similarity of $0.23$ (caused by named entities, numbers and punctuation which can remain unchanged).
On the opposite end, Reference had the highest similarity followed by Google, M11 and M07.
The last two columns and linear fits are discussed in \Cref{subsec:quality}.

\begin{table}[ht]
    \center
    \begin{tabular}{lccc}
\toprule
\textbf{Model} & \textbf{P0$\rightarrow$P1} & \textbf{P1$\rightarrow$P2} & \textbf{P0$\rightarrow$P2} \\
\midrule
Source & 0.23 & 0.88 & 0.23 \\
M01 & 0.65 & 0.94 & 0.63 \\
M02 & 0.75 & 0.92 & 0.71 \\
M03 & 0.72 & 0.90 & 0.69 \\
M04 & 0.74 & 0.88 & 0.70 \\
M05 & 0.74 & 0.94 & 0.73 \\
M06 & 0.77 & 0.93 & 0.74 \\
M07 & 0.80 & 0.93 & 0.78 \\
M08 & 0.77 & 0.94 & 0.76 \\
M09 & 0.77 & 0.93 & 0.76 \\
M10 & 0.77 & 0.94 & 0.77 \\
M11 & 0.80 & 0.95 & 0.80 \\
M11* & - & - & 0.92 \\
Google & 0.80 & 0.93 & 0.76 \\
Microsoft & 0.74 & 0.91 & 0.70 \\
Reference & 0.90 & 0.96 & 0.87 \\
Reference* & - & - & 0.87 \\
\midrule
\textbf{Average}  & 0.73 & 0.93 & 0.73 \\
\midrule
\textbf{Lin. fit, all} & 0.011 & 0.001 & 0.015 \\
\textbf{Lin. fit, $>$36} & 0.004 & 0.000 & 0.027 \\
\bottomrule
\end{tabular}
    \caption{Average ChrF similarity per system between different stages of post-editing. Bottom two lines show linear fit coefficient on either all MT systems or on MT systems with BLEU $>36$ (reference and source excluded). P0: system output, P1: post-editors' output, P2: reviewers' output.}
    \label{tab:axp3}
\end{table}

The post-editing of Reference had on average ChrF of $0.90$, while MT models $0.75\pm0.04$. An improvement in one BLEU point then corresponds to $1.47\%$ of the MT average ChrF.

\begin{figure}[ht]
    \center
    \includegraphics[width=\linewidth]{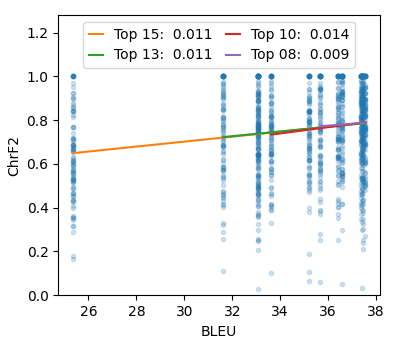}
    \caption{Sentence similarity measured by ChrF2 between the provided translation and first-phase (P0$\rightarrow$P1). Every dot is a single sentence translated by a given MT. Source and Reference measurements are omitted for scale.}
    \label{fig:exp5}
\end{figure}

\Cref{fig:exp5} shows the trend of the relationship of MT quality and post-edited distance (first phase, P0$\rightarrow$P1) measured by ChrF2. It is systematically positive even when considering only top-n MT systems. Graphs for P1$\rightarrow$P2 and P0$\rightarrow$P2 (not shown) suggest that there is also a small correlation between the MT systems' quality and post-edited distance for the second phase (P1$\rightarrow$P2).
This trend is similar when using TER instead of BLEU and the figure is shown in \Cref{app_sec:exp5_ter}.

\paragraph{Unigram Comparison}

To examine the proportion of words that were only moved within the sentence, we also computed unigram precision and recall between the provided translations and the post-edited outputs. As expected, most of the words in the reference translation were unchanged by the post-editors
($F_1 = 0.92$), this is in contrast to the MT systems ($F_1 = 0.78 \pm 0.04$).

In this case, the linear relationship to BLEU is preserved from all models (slope $0.011$) to only the top eight models (slope $0.008$). The first corresponds to $1.41\%$.

\subsection{Time}\label{sec:time}

% We wish to also include more concrete examples that show why this is necessary.

We focus on time spent per one token, which is more useful in determining the overall time a post-editor has to spend working with a document than time spent per one sentence.
The CAT tool records two quantities for each segment:
\begin{itemize}
\item Think time: the time between entering a given segment (i.e. the keyboard cursor moves to that segment) and doing the first edit operation
\item Edit time: the sum of thinking time and the time spent editing the segment (until the translation is finished and confirmed)
\end{itemize}
However, both of these measured quantities carry some level of noise due to translator breaks and other distractions in the think time.

The post-editors were instructed to take breaks only at document boundaries, but there were a number of deviations\footnote{Maximum time per word was 1482s, which is highly improbable.} which can be explained only by the post-editor getting distracted by other activities. Most of these deviations were present already in think time. Let $T$ and $W$ be the true variables for think and write times per word and $\hat{T}$ and $\hat{A}$ (all time) our measured estimates. The term $\epsilon_T$ is then causing the high deviation in $\hat{T}$ and subsequently in $\hat{A}$.

\begin{align*}
\hat{T} &\approx T + \epsilon_T & \text{Measured think time} \\
\hat{A} &\approx \hat{T} + \hat{W} & \text{Measured total time}\\
&= T + W + \epsilon_T + \epsilon_W & \\
\overset{*}{W} &\coloneqq \hat{A} - \hat{T} & \text{Measured write time} \\
&\approx W + \epsilon_W & \\
\overset{*}{T} &\coloneqq \min\{10s, \hat{T}\} & \text{Estimated think time} \\
\overset{*}{A} &\coloneqq \overset{*}{W} + \min\{10s, \hat{T}\} & \text{Estimated total time}
\end{align*}

The two quantities $\overset{*}{T}$ and $\overset{*}{A}$ then approximate the think time and total time, respectively. The latter is used in the following figures and referred to as estimated total time. Think time is estimated by capping the value per word to 10s.\footnote{We chose 10 seconds to cap the think time per word because it seemed improbable that anyone would genuinely spend all this time thinking about the upcoming sentence.}
The choice of filtering has a significant impact on the result. Even though the current strategy was chosen with the best intentions, it is unclear whether it is universally the optimal one.
The interest in the variable of the total time is sparked by the immediate commercial relevance: \textit{How does one BLEU point in used MT system affect the total work time of post-editors spent on one word?}\footnote{Prices of translations are usually calculated by the number of words in the document.}

\Cref{tab:txp1} shows the estimates for think and total times with 95\% confidence intervals. Although there is some overlap between the systems, they are spread out evenly between 6s and 12s. Reference ranked by far the first and Source in the middle. For all systems below Source, post-editing MT output took longer than translating from scratch.

\begin{table}[ht]
    \center
    \begin{tabular}{lcc}
\toprule
\textbf{Model} & \textbf{Total time} & \textbf{Think time} \\
\midrule
 Reference &   3.17s$\pm$0.13s &   0.58s$\pm$0.04s \\
       M08 &   4.10s$\pm$0.20s &   0.55s$\pm$0.03s \\
    Google &   4.52s$\pm$0.22s &   0.96s$\pm$0.08s \\
       M03 &   4.60s$\pm$0.19s &   0.60s$\pm$0.04s \\
       M07 &   4.95s$\pm$0.27s &   0.92s$\pm$0.06s \\
       M01 &   5.13s$\pm$0.18s &   0.97s$\pm$0.05s \\
       M09 &   5.41s$\pm$0.36s &   1.12s$\pm$0.07s \\
       M05 &   5.64s$\pm$0.21s &   0.93s$\pm$0.07s \\
    Source &   6.00s$\pm$0.22s &   0.72s$\pm$0.05s \\
 Microsoft &   6.02s$\pm$0.32s &   0.87s$\pm$0.06s \\
       M04 &   6.27s$\pm$0.27s &   1.46s$\pm$0.09s \\
       M02 &   6.44s$\pm$0.27s &   1.16s$\pm$0.07s \\
       M10 &   6.45s$\pm$0.32s &   2.31s$\pm$0.12s \\
       M11 &   8.01s$\pm$0.47s &   1.63s$\pm$0.09s \\
       M06 &   8.25s$\pm$0.39s &   1.62s$\pm$0.07s \\
\midrule
\textbf{Average} &  5.66s$\pm$0.07s &   1.09s$\pm$0.02s \\
\bottomrule
\end{tabular}
    \caption{Total and think time estimations for first phase of post-editing for all MT systems (+Source and Reference). Confidence intervals computed for 95\%. Sorted by total time.}
    \label{tab:txp1}
\end{table}

\begin{figure}[ht]
    \center
    \includegraphics[width=\linewidth]{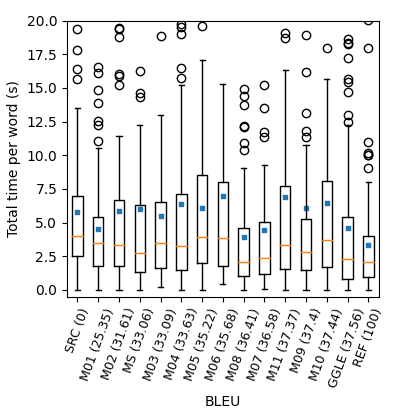}
    \caption{Total time per word in relation to MT system BLEU score. Every dot is a single post-edited sentence.
    % Legend shows slopes of fitted linear functions on top-k systems. Source and Reference measurements are omitted for scale.
    Zoomed to [0, 20] on the y-axis. Orange bars represent medians and blue squares means. Upper whiskers are the 3rd quartile + $1.5\times$ inter-quartile range.
    }
    \label{fig:exp9}
\end{figure}

The relationship between BLEU and total time per word is shown in \Cref{fig:exp9}. It shows no clear systematic relationship between the two variables.
% The graph shows linear least-squares polynomial fit. We provide measurements for top-k systems because (1) the worse systems receive higher individual weight since they are further away from the cluster, and (2) we are mostly interested in high-performance systems.
For comparison with \citet{sanchez2016machine} we also report slopes of linear least-squares fits.
A slope of $0.044$ (all MT systems)
% represents a relationship that corresponds to 
indicates that a 1 BLEU point increase in MT quality increases total time per word by $0.044$s (i.e., that higher-quality MT may in fact lead to slightly longer post-editing).
For top-8 systems, the slope is negative, meaning that a 1 BLEU point increase decreases total word time by $0.514$s.
However, these results should not be interpreted in the sense that for high-quality MT systems, BLEU improvements lead to faster post-editing. On the contrary, they should illustrate the uncertainty and complexity of the relationship between the two quantities.

% Ondra: V nasledujicim vypoctu je velky problem, ze Source nema 0 a Reference nema 100. Skoro bych to radsi vynechal.
% Adding Reference and Source with BLEU scores of 100 and 0 respectively yields a slope of $-0.030$ (similar for just Reference or just Source with the rest of the MT systems).

\subsection{Quality} \label{subsec:quality}

% To measure the quality of the post-editors' output, a set of reviewers was used.
The quality of post-editors' output was measured during revision.
This closely follows industry standards, where the text to translate is first given to post-editors and then to another set of reviewers. Here we again used ChrF to determine how much effort was needed to create production-level translations from the already post-edited translations.

Apart from the outputs of the work of the first phase of post-editors, we also mixed in unedited Reference (labelled Reference*) and M11 (M11*).
% This way the effect of post-editing can be quantitatively observed.
This allows us to see if there is any effect of priming the workers with the task specification: post-editors are likely to expect to have more work with fixing MT output than reviewers.

In this case, the total and think times were estimated the same way as for the first phase. The results per system are shown in \Cref{tab:txp2}. The distribution is now more uniform, and in comparison to the first phase, shown in \Cref{tab:txp1}, many systems changed their rank. Documents of M11* and Reference* (not post-edited in the first phase) had much larger average total times than their post-edited versions, M11 and Reference. This is caused by more thorough reviewing necessary since the documents were not refined. Furthermore, the reviewers may not have expected this kind of input at all. Note however that the total time for M11* is still not much higher than the average time required to review an already post-edited MT output. % Ales sem pridal jednu vetu "pro zajimavost"
% Supr -vz
% \OB{Napsat duvod? We can speculate that this is because the reviewers were generally not expecting any problems but ran into them and a forced change of strategy took some extra time.}

% Přepsal jsem to ovšem asi bude lepší to smazat.
% Ales souhlasi se smazanim a maze :-)
% Translation from scratch (Source) has a similar average total review time ($3.37$) in phase two as does Reference in the first phase ($3.17$).
%Although not strictly comparable across the two phases (different setup), it is a check which confirms that the reference translation created independently is of roughly the same quality as the translations created from scratch in our experiment.

\begin{table}[ht!]
    \center
    \begin{tabular}{lcc}
\toprule
\textbf{Model} & \textbf{Total time} & \textbf{Think time} \\
\midrule
       M08 &   2.12s$\pm$0.11s &   0.96s$\pm$0.07s \\
       M01 &   2.29s$\pm$0.14s &   0.96s$\pm$0.06s \\
 Reference &   2.32s$\pm$0.12s &   0.97s$\pm$0.06s \\
       M11 &   2.34s$\pm$0.11s &   1.10s$\pm$0.06s \\
       M06 &   2.53s$\pm$0.17s &   0.96s$\pm$0.05s \\
       M02 &   2.98s$\pm$0.18s &   0.83s$\pm$0.04s \\
    Google &   3.12s$\pm$0.13s &   1.31s$\pm$0.07s \\
       M07 &   3.36s$\pm$0.22s &   1.19s$\pm$0.08s \\
    Source &   3.37s$\pm$0.12s &   1.01s$\pm$0.05s \\
       M04 &   3.70s$\pm$0.13s &   1.10s$\pm$0.06s \\
       M05 &   3.75s$\pm$0.28s &   1.05s$\pm$0.06s \\
 Microsoft &   3.75s$\pm$0.22s &   1.12s$\pm$0.06s \\
      M11* &   3.96s$\pm$0.30s &   1.17s$\pm$0.08s \\
       M03 &   4.06s$\pm$0.16s &   0.87s$\pm$0.05s \\
       M09 &   4.41s$\pm$0.23s &   0.85s$\pm$0.06s \\
       M10 &   4.83s$\pm$0.31s &   1.71s$\pm$0.08s \\
Reference* &   5.31s$\pm$0.18s &   1.52s$\pm$0.07s \\
\midrule
\textbf{Average} &  3.42s$\pm$0.05s &   1.10s$\pm$0.02s \\
\bottomrule
\end{tabular}
    \caption{Total and think time estimations for the review phase of post-editing for all MT systems (+Source and Reference). Confidence intervals computed for 95\%. Sorted by total time.}
    \label{tab:txp2}
\end{table}

\begin{table}[ht!]
    \renewcommand{\arraystretch}{1.0}
    \centering
    \begin{tabular}{lcccc}
        \toprule
        \hspace{-0.2cm} Model/Doc. \hspace{-1cm}  & Acc. & Flu. & Other & All \\
        \midrule
Source & \blocksimple{ 0.20202020202020202 } & \blocksimple{ 0.2727272727272727 } & \blocksimple{ 0.13131313131313133 } & \blocksimple{ 0.6060606060606061 } \\
M01 & \blocksimple{ 0.16161616161616163 } & \blocksimple{ 0.1111111111111111 } & \blocksimple{ 0.020202020202020204 } & \blocksimple{ 0.29292929292929293 } \\
M02 & \blocksimple{ 0.1919191919191919 } & \blocksimple{ 0.1919191919191919 } & \blocksimple{ 0.10101010101010101 } & \blocksimple{ 0.48484848484848486 } \\
M03 & \blocksimple{ 0.3838383838383838 } & \blocksimple{ 0.15151515151515152 } & \blocksimple{ 0.06060606060606061 } & \blocksimple{ 0.5959595959595959 } \\
M04 & \blocksimple{ 0.3434343434343434 } & \blocksimple{ 0.1919191919191919 } & \blocksimple{ 0.010101010101010102 } & \blocksimple{ 0.5454545454545454 } \\
M05 & \blocksimple{ 0.20202020202020202 } & \blocksimple{ 0.1717171717171717 } & \blocksimple{ 0.030303030303030304 } & \blocksimple{ 0.40404040404040403 } \\
M06 & \blocksimple{ 0.08247422680412371 } & \blocksimple{ 0.15463917525773196 } & \blocksimple{ 0.07216494845360824 } & \blocksimple{ 0.30927835051546393 } \\
M07 & \blocksimple{ 0.3434343434343434 } & \blocksimple{ 0.1414141414141414 } & \blocksimple{ 0.06060606060606061 } & \blocksimple{ 0.5454545454545454 } \\
M08 & \blocksimple{ 0.2727272727272727 } & \blocksimple{ 0.15151515151515152 } & \blocksimple{ 0.0 } & \blocksimple{ 0.42424242424242425 } \\
M09 & \blocksimple{ 0.43434343434343436 } & \blocksimple{ 0.1717171717171717 } & \blocksimple{ 0.020202020202020204 } & \blocksimple{ 0.6262626262626263 } \\
M10 & \blocksimple{ 0.21428571428571427 } & \blocksimple{ 0.12244897959183673 } & \blocksimple{ 0.030612244897959183 } & \blocksimple{ 0.3673469387755102 } \\
M11 & \blocksimple{ 0.1919191919191919 } & \blocksimple{ 0.09090909090909091 } & \blocksimple{ 0.0 } & \blocksimple{ 0.2828282828282828 } \\
M11* & \blocksimple{ 0.6060606060606061 } & \blocksimple{ 0.1919191919191919 } & \blocksimple{ 0.030303030303030304 } & \blocksimple{ 0.8282828282828283 } \\
Google & \blocksimple{ 0.16161616161616163 } & \blocksimple{ 0.2828282828282828 } & \blocksimple{ 0.0 } & \blocksimple{ 0.4444444444444444 } \\
Microsoft & \blocksimple{ 0.2222222222222222 } & \blocksimple{ 0.2727272727272727 } & \blocksimple{ 0.010101010101010102 } & \blocksimple{ 0.5050505050505051 } \\
Reference & \blocksimple{ 0.20202020202020202 } & \blocksimple{ 0.13131313131313133 } & \blocksimple{ 0.010101010101010102 } & \blocksimple{ 0.3434343434343434 } \\
Reference* & \blocksimple{ 0.3838383838383838 } & \blocksimple{ 0.13131313131313133 } & \blocksimple{ 0.0707070707070707 } & \blocksimple{ 0.5858585858585859 } \\

\midrule

News & \blocksimple{ 0.3937908496732026 } & \blocksimple{ 0.30718954248366015 } & \blocksimple{ 0.03758169934640523 } & \blocksimple{ 0.738562091503268 } \\
Audit & \blocksimple{ 0.25831202046035806 } & \blocksimple{ 0.12020460358056266 } & \blocksimple{ 0.023017902813299233 } & \blocksimple{ 0.40153452685422 } \\
Technical & \blocksimple{ 0.19786096256684493 } & \blocksimple{ 0.1497326203208556 } & \blocksimple{ 0.053475935828877004 } & \blocksimple{ 0.40106951871657753 } \\
Lease & \blocksimple{ 0.15510204081632653 } & \blocksimple{ 0.05510204081632653 } & \blocksimple{ 0.04693877551020408 } & \blocksimple{ 0.2571428571428571 } \\
    \end{tabular}
    
    \caption{Average LQA severity (reported from 0 to 3) of models and documents across three categories: Adequacy/accuracy, fluency and other. Their average is reported in the last column. Empty and full squares represent severities of 0 and 1, respectively.}
    \label{tab:axp5}
    \vspace{-0.5cm}
\end{table}

In contrast to \Cref{fig:exp9}, the linear least-square fit slope for total times of top-15 and top-8 are $0.069$ and $0.765$ in the case of reviewing.
This suggests that an improvement in BLEU may lead to higher times when reviewing the post-edited output and that BLEU may be not a good predictor of overall localization effort. We currently do not have an explanation for this effect.

% An improvement in BLEU leads to higher times in the work of reviewers working on already post-edited output.
% This clearly documents that BLEU is not a good predictor of overall localization effort.

The reviewers were also tasked to mark errors using LQA. For every sentence, we sum the LQA severities and compute an average for every model.
There was no significant linear relationship between the average severity and overall performance measured by BLEU.
Exceptions (deviating from the average $0.51$)
are Reference* ($0.59$) and M11* ($0.83$), which were not post-edited in the first phase. 
Source had an average severity of $0.61$, while the best system, M11, had the lowest $0.28$.
There was, however, a significant difference between the average severity of documents: Lease ($0.26$), Audit ($0.40$), Technical ($0.40$) and News ($0.74$).
The average LQA severity is shown in \Cref{tab:axp5}.

\paragraph{Output Similarity}

\Cref{tab:axp3} shows the similarities between the output of the first phase and the second phase (second column) and the system output and the second phase (third column). For M11* and Reference*, the output of the first phase is undefined. The similarities in the second column are much more dispersed, though still Reference was post-edited the less ($0.96$, while Source the most ($0.88$). A similar thing can be observed between the system output and final output, with the exception of non-post-edited M11* being post-edited very little. In fact, the raw MT output was modified less than some of the already post-edited translations. Reference post-edited by first phase post-editors had a similarity to the original $0.90$, which is very similar to Reference post-edited only by the reviewer (Reference*, $0.87$).

Linear functions are fitted to see the effect of one BLEU point on the amount of post-editing (measured by ChrF). The results are in the bottommost lines of \Cref{tab:axp3}. The effect is the strongest when measuring the similarity between the model output and the second phase. The linear fit is, however, strongly influenced by the less-performing models. In the case where only the top eight models (BLEU $> 36$) are taken, an increase of 1 BLEU point corresponds to $0.027$ increase in similarity between model output and final version of the sentence ($\sim\hspace{-0.08cm}3.7\%$ of the total average).
A similar trend (negative slope) was observed also when using TER instead of BLEU.
Reference and Source were excluded from this computation because their artificial BLEU scores (100 and 0 respectively) would have an undesired effect on the result.
The average similarity between the once post-edited output and the corrections is $0.93$, confirming the hypothesis that most of the errors are resolved in the first pass of post-editing.

\paragraph{Editing process.}

\Cref{tab:txp3} shows the breakdown of edit types\footnote{Using the Ratcliff-Obershelp algorithm \citep{ratcliff1988pattern} implemented in Python \texttt{difflib}.} between the provided translation and the post-edited output. The \textit{insert} and \textit{delete} values show no apparent trend, though \textit{replace} is decreasing with higher BLEU scores.

\begin{table}[ht!]
    \center
    \begin{tabular}{lccc}
\toprule
\textbf{Model} & \textbf{Replace} & \textbf{Delete} & \textbf{Insert} \\
\midrule
    Source & $ 22.22$ & $  0.00$ & $  0.02$  \\
       M01 & $  9.71$ & $  0.64$ & $  0.69$  \\
       M02 & $  6.84$ & $  0.30$ & $  0.66$  \\
       M03 & $  7.48$ & $  0.51$ & $  0.71$  \\
       M04 & $  6.49$ & $  0.41$ & $  0.54$  \\
       M05 & $  6.67$ & $  0.48$ & $  0.40$  \\
       M06 & $  6.32$ & $  0.78$ & $  0.59$  \\
       M07 & $  5.49$ & $  0.33$ & $  0.62$  \\
       M08 & $  6.03$ & $  0.46$ & $  0.38$  \\
       M09 & $  5.84$ & $  0.32$ & $  0.52$  \\
       M10 & $  5.70$ & $  0.61$ & $  0.68$  \\
       M11 & $  5.62$ & $  0.40$ & $  0.94$  \\
    Google & $  5.32$ & $  0.32$ & $  0.44$  \\
 Microsoft & $  7.29$ & $  0.42$ & $  0.98$  \\
 Reference & $  2.96$ & $  0.18$ & $  0.42$  \\
\midrule
\textbf{Average} &$  7.34$ & $  0.41$ & $  0.57$  \\
\bottomrule
\end{tabular}
    \caption{Average number of line edit operations for the first phase of post-editing for all MT systems (+Source and Reference). For specific operations, \textbf{Insert} considers the number of target tokens, \textbf{Delete} the number of source tokens and \textbf{Replace} their average.}
    \label{tab:txp3}
\end{table}
 
\section{Summary}

In this work, we extended the standard scenario for testing post-editing productivity by a second phase of annotations. This allowed for further insight into the quality of the output and it also follows the standard two-phase process of translation companies more closely.

% This answers the initial questions

% Q: How strong is the relationship between MT quality and post-editing speed?
% We found a complex relationship between MT quality and post-editing speed, which depends on whether only state-of-the-art systems are considered or also artificially worse ones. For the top 8 systems, an improvement of one BLEU point corresponded to $0.514$ fewer seconds per one word on average.

We found a complex relationship between MT quality and post-editing speed, which depends on many factors. When considering only the top 8 systems, an improvement of one BLEU point corresponded to $0.514$ fewer seconds per one word on average but at the same time, this trend was not confirmed on larger sets of systems. Overall, the relationship is most likely weaker than previously assumed.

% 
% Q: Does MT quality have a measurable impact on the quality of the post-edited translation?
We did not find any significant relationship between the produced output quality and MT system performance among all systems because
% 
% Q: Is the effect of MT quality still persistent in the second phase of post-editing?
the effect was not measurable
in the second phase.
% on the second phase productivity.
% 
% Q: Is the post-editing process any different when the input is the reference itself?
As expected, post-editing human reference led to the smallest amount of edits and time spent.
Contrary to current results, translating from scratch was not significantly slower than post-editing in either of the two phases.
% 
% Q: How large are the edits in the first and the second phase of annotation?
The average ChrF similarity between the provided output and the first phase results was $0.73$ and between the two phases $0.93$, suggesting diminishing results of additional phases.

% \vspace{-0.2cm}    
% \paragraph{Future work}

% Although this was not the focus of this work, we observed high variance between both the domains and the participants. It remains unclear as to whether one causes the other or whether they are unrelated.

The most significant conclusion is that for NMT, the previously assumed
%linking hypothesis
link
between MT quality and post-editing time is weak and not straightforward.
The current recommendation for the industry is that they should not expect small improvements in MT (measured by automatic metrics) to lead to significantly lower post-editing times nor significantly higher post-edited quality.

\section*{Ethics}

Both post-editors and proofreaders were compensated by their usual professional wages.

\section*{Acknowledgments}
% We sincerely thank České překlady for offering collaboration on the experiment and the company Memsource, from which this project has received funding for the annotators.
We sincerely thank České překlady for their collaboration and for providing all translations and revisions.
The work was supported by Memsource and by the grants
 19-26934X (NEUREM3) and % Ondřej
 20-16819X (LUSyD) by the Czech Science Foundation. % Martin

%We also used language resources developed, stored and distributed by the LINDAT-Clarin project of the Ministry of Education of the Czech Republic (project LM2010013).
The work has been using language resources developed and distributed by the
 LINDAT/CLARIAHCZ project of the Ministry of Education, Youth
 and Sports of the Czech Republic (project LM2018101).

\bibliography{misc/mybib}
\bibliographystyle{misc/acl_natbib}

%
% File naacl2019.tex
%
%% Based on the style files for ACL 2018 and NAACL 2018, which were
%% Based on the style files for ACL-2015, with some improvements
%%  taken from the NAACL-2016 style
%% Based on the style files for ACL-2014, which were, in turn,
%% based on ACL-2013, ACL-2012, ACL-2011, ACL-2010, ACL-IJCNLP-2009,
%% EACL-2009, IJCNLP-2008...
%% Based on the style files for EACL 2006 by 
%%e.agirre@ehu.es or Sergi.Balari@uab.es
%% and that of ACL 08 by Joakim Nivre and Noah Smith

% \documentclass[11pt,a4paper]{article}
% \usepackage{emnlp2021}
% \usepackage{times}
% \usepackage{latexsym}
% \usepackage{enumitem}
% \usepackage{booktabs}
% \usepackage{array}
% \usepackage{url}
% \usepackage{tikz}
% \usepackage{amsmath}
% \usepackage{mathtools}
% \usepackage{cleveref}
% \usepackage{ulem}
% \graphicspath{{img/}}

% \input{macros.tex}

% \date{}

% \begin{document}
% \maketitle

\newpage 

\appendix

\section{Questionare for Second-Phase} \label{sec:questionnaire_p2}

\begin{table}[ht]
    \center
    \begin{tabular}{lccc}
\toprule
(years) & $<$ 5 & $<$ 10 & $\ge$ 10 \\
\midrule
Translation experience\hspace{0.0cm} & 1 & 2 & 14 \\
Post-Editor experience\hspace{0.0cm} & 10 & 3 & 4 \\
\bottomrule
\end{tabular}
    \caption{Reviewers' (second phase) years of translation and post-editing experience.}
    \label{tab:experience_p2}
\end{table}

\begin{table}[ht]
    \center
    \begin{tabular}{>{\raggedright}p{5.7cm}m{1.15cm}}
\toprule
Question &\hspace{-0.35cm}Response\hspace{-0.2cm} \\
\midrule
Comfortable with post-editing human-like (perfect) quality & \blockfirst{1}{2}{6}{3}{5} \\
Comfortable with post-editing less-than-perfect quality & \blockfirst{1}{7}{5}{3}{1} \\
Prefer PE to translating from scratch (without a TM) & \blockfirst{1}{12}{2}{1}{1} \\
MT helps to maintain translation consistency & \blockfirst{3}{6}{5}{3}{3} \\
MT helps to translate faster & \blockfirst{5}{3}{0}{9}{0} \\
PE is more laborious than translating from scratch or with a TM & \vspace{0.4cm} \blockfirst{1}{2}{5}{7}{2} \\
Prefer PE to processing 85--94\% TM matches & \blockfirst{3}{8}{6}{0}{0} \\
Prefer PE to editing a human translation & \blockfirst{0}{7}{10}{0}{0} \\
% new questions
MT helps to produce better results & \blockfirst{3}{3}{4}{7}{0} \\
Often use MT outside of work for known languages & \blockfirst{3}{7}{5}{2}{0} \\
Often use MT outside of work for unknown languages & \blockfirst{1}{2}{3}{8}{3} \\
\bottomrule
\end{tabular}
    \caption{Reviewers' (second phase) answers regarding their profession on the Likert scale (leftmost bar = Strongly Disagree, rightmost bar = Strongly Agree), TM = translation memory.}
    \label{tab:questionare_p2}
\end{table}

\vspace{4cm}

\section{TER as an Evaluation Measure} \label{app_sec:exp5_ter}

\begin{figure}[ht]
    \center
    \includegraphics[width=\linewidth]{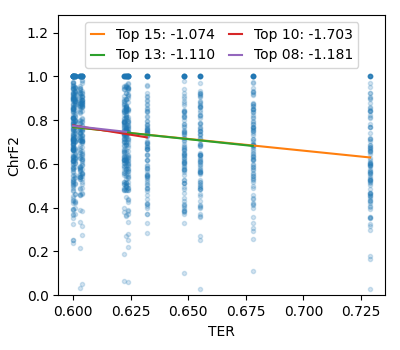}
    \caption{Sentence similarity measured by ChrF2 between the provided translation and first-phase (P0$\rightarrow$P1) in contrast to system TER score (lower is better). Every dot is a single sentence translated by a given MT. Source and Reference measurements are omitted for scale.}
\end{figure}

\end{document}